\documentclass[11pt]{article}

\usepackage[preprint]{acl}

\usepackage{amsmath,amsfonts,bm}

\def\eqref#1{equation~\ref{#1}}

\def\1{\bm{1}}

\DeclareMathAlphabet{\mathsfit}{\encodingdefault}{\sfdefault}{m}{sl}
\SetMathAlphabet{\mathsfit}{bold}{\encodingdefault}{\sfdefault}{bx}{n}

\DeclareMathOperator*{\argmax}{arg\,max}

\usepackage{times}
\usepackage{latexsym}
\usepackage[T1]{fontenc}
\usepackage[utf8]{inputenc}
\usepackage{microtype}
\usepackage{inconsolata}

\usepackage{graphicx}
\usepackage{booktabs}
\usepackage{multirow}
\usepackage{bbm}
\usepackage{algorithm}
\usepackage{algorithmic}
\usepackage[table,svgnames]{xcolor}
\usepackage{enumitem}
\usepackage[most]{tcolorbox}

\definecolor{promptbg}{RGB}{245, 245, 250}
\definecolor{promptborder}{RGB}{100, 100, 180}
\newtcolorbox{promptbox}[1][]{%
    enhanced,
    colback=promptbg,
    colframe=promptborder,
    fonttitle=\bfseries,
    title=#1,
    left=5pt,
    right=5pt,
    top=5pt,
    bottom=5pt,
}

\graphicspath{{figures/}}

\newcommand{\methodName}{\textsc{Step-Grm}}

\title{Making Evaluation Explicit Is Not Enough: \\ Supervising the Evaluation Process in Generative Reward Models}

\author{\textbf{Qiyuan Chen}\textsuperscript{1,2},
  \textbf{Yuze Zhao}\textsuperscript{2},
  \textbf{Jiahe Chen}\textsuperscript{1},
  \textbf{Hongsen Huang}\textsuperscript{3},\\
  \textbf{Hongxia Xu}\textsuperscript{1},
  \textbf{Yingda Chen}\textsuperscript{2}\thanks{Corresponding author.},
  \textbf{Jian Wu}\textsuperscript{1}\textsuperscript{$\ast$} \\
  \textsuperscript{1}Zhejiang University, \textsuperscript{2}Token Foundry, Alibaba Group, \textsuperscript{3}Soochow Securities Co.,Ltd. \\
  \texttt{chenqiyuan1012@gmail.com}}

\begin{document}
\maketitle

\begin{abstract}
Generative reward models (GRMs) make language-model evaluation more explicit by generating critiques, rubrics, or intermediate analyses before predicting a final preference. However, explicit evaluation does not necessarily imply a reliable evaluation process: under outcome-only supervision, a model may reach the correct final preference while producing weak or unsupported intermediate judgments, turning the evaluation into a post-hoc rationalization.
We introduce \methodName{}, a framework that directly supervises the intermediate evaluation process of rubric-based GRMs. In addition to conventional preference supervision, \methodName{} provides structural guidance for rubric construction and local supervision for criterion-level judgments, encouraging the model to learn not only the correct final preference but also a more reliable process for reaching it.
Extensive experiments show that \methodName{} achieves strong performance on standard reward-modeling benchmarks while improving the consistency and correctness of intermediate evaluations. Moreover, reward signals produced by \methodName{} lead to better downstream policy optimization, demonstrating that more reliable evaluation processes can translate into practical gains beyond reward-model accuracy.
\end{abstract}

\section{Introduction}
\label{sec:introduction}

\begin{figure}[t]
\centering
\includegraphics[width=\columnwidth]{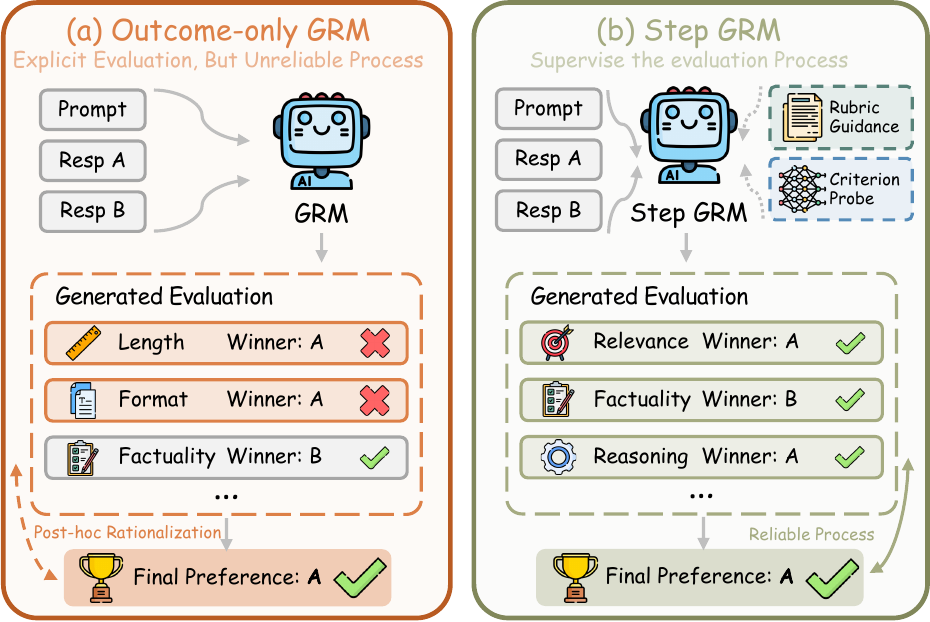}
\caption{\textbf{Making evaluation explicit is not enough.} Outcome-only supervision can yield the correct final preference while leaving intermediate criterion-level judgments unsupported, allowing the generated evaluation to become a post-hoc rationalization. \methodName{} supervises both rubric construction and criterion-level judgment, turning the evaluation process itself into a direct object of learning.}
\label{fig:teaser}
\vspace{-2mm}
\end{figure}


Reward models are fundamental to aligning large language models, providing supervision for reinforcement learning and automated evaluation~\citep{ouyang2022training,bai2022constitutional,zheng2023judging,rafailov2023direct}. Conventional reward models typically map candidate responses directly to a single scalar score or preference label, compressing complex, multi-dimensional judgments into an opaque prediction~\citep{wang2024interpretable,wang2026helpsteer3,liu2026skywork}. To improve transparency, generative reward models (GRMs) have emerged as an expressive alternative that produces explicit reasoning traces, such as critiques or rubrics, before predicting the final preference~\citep{kim2024prometheus,zhang2025generative}.

Recent work increasingly uses reinforcement learning to train generative reward models, encouraging them to produce dynamic rubrics before predicting preferences~\citep{yuan2024self,chen2026rm,huang2026think}. However, outcome-only rewards verify whether the final winner is correct, while leaving the intermediate reasoning unconstrained. Driven purely by this final reward, the model can rely on superficial cues such as response length or style to determine the winner, and then craft intermediate judgments that merely support that choice. Instead of serving as independent evidence, the generated rubric becomes a post-hoc rationalization of the predetermined outcome~\citep{lai2026r,wang2026outcome}. As shown in Figure~\ref{fig:teaser}, making the evaluation explicit is therefore not enough; the intermediate judgments themselves must also be supervised.


This motivates a fundamental question: \textbf{What must a reliable GRM actually get right?} At a minimum, sound evaluation requires three sequential decisions: identifying which criteria matter, comparing the responses under each individual criterion, and integrating these criterion-level judgments into a final preference. A final preference label alone supervises only the last step, leaving the first two decisions entirely unguided. Bridging this gap requires dedicated supervision of intermediate evaluation: reference rubrics to teach the model which criteria to examine, and independent criterion-level checks to ensure that these judgments are factually supported.


We introduce \methodName{} (\textbf{S}upervising \textbf{T}he \textbf{E}valuation \textbf{P}rocess in \textbf{G}enerative \textbf{R}eward \textbf{M}odels), a framework that complements outcome rewards with two process supervision signals during reinforcement learning. First, we provide privileged rubric guidance: high-quality reference rubrics are treated as privileged training context to guide rubric construction through self-distillation, teaching the model to construct effective criteria without requiring references at inference time. Second, we obtain criterion-level judgments using a frozen, single-criterion probe. This probe independently judges the responses under each policy-generated criterion, preventing local verdicts from blindly echoing the final preference. Both the reference rubrics and the probe are used exclusively during training, allowing \methodName{} to operate fully autonomously at inference time.

Experiments on \textsc{RewardBench} and \textsc{RM-Bench} show that \methodName{} consistently outperforms competitive scalar and generative baselines in preference discrimination. On \textsc{HelpSteer3}, our framework substantially improves Process Consistency and Local Accuracy relative to outcome-only training, effectively suppressing rationalization drift~\citep{wang2025helpsteer3}. Furthermore, guiding downstream preference optimization via DPO with \methodName{} achieves higher win rates on \textsc{Arena-Hard} and \textsc{AlpacaEval}, confirming that a reliable evaluation process translates into superior generation quality.

In summary, our contributions are threefold: (1) we identify the intermediate supervision gap in generative reward modeling, showing that outcome-only reinforcement learning induces post-hoc rationalization; (2) we propose \methodName{}, which directly supervises rubric construction and criterion-level judgment through privileged rubric self-distillation and a frozen single-criterion probe; and (3) we empirically validate across diverse benchmarks that supervising the evaluation process simultaneously enhances preference discrimination, intermediate faithfulness, and downstream policy alignment.
\section{Method}
\label{sec:method}

\begin{figure*}[t]
\centering
\includegraphics[width=\textwidth]{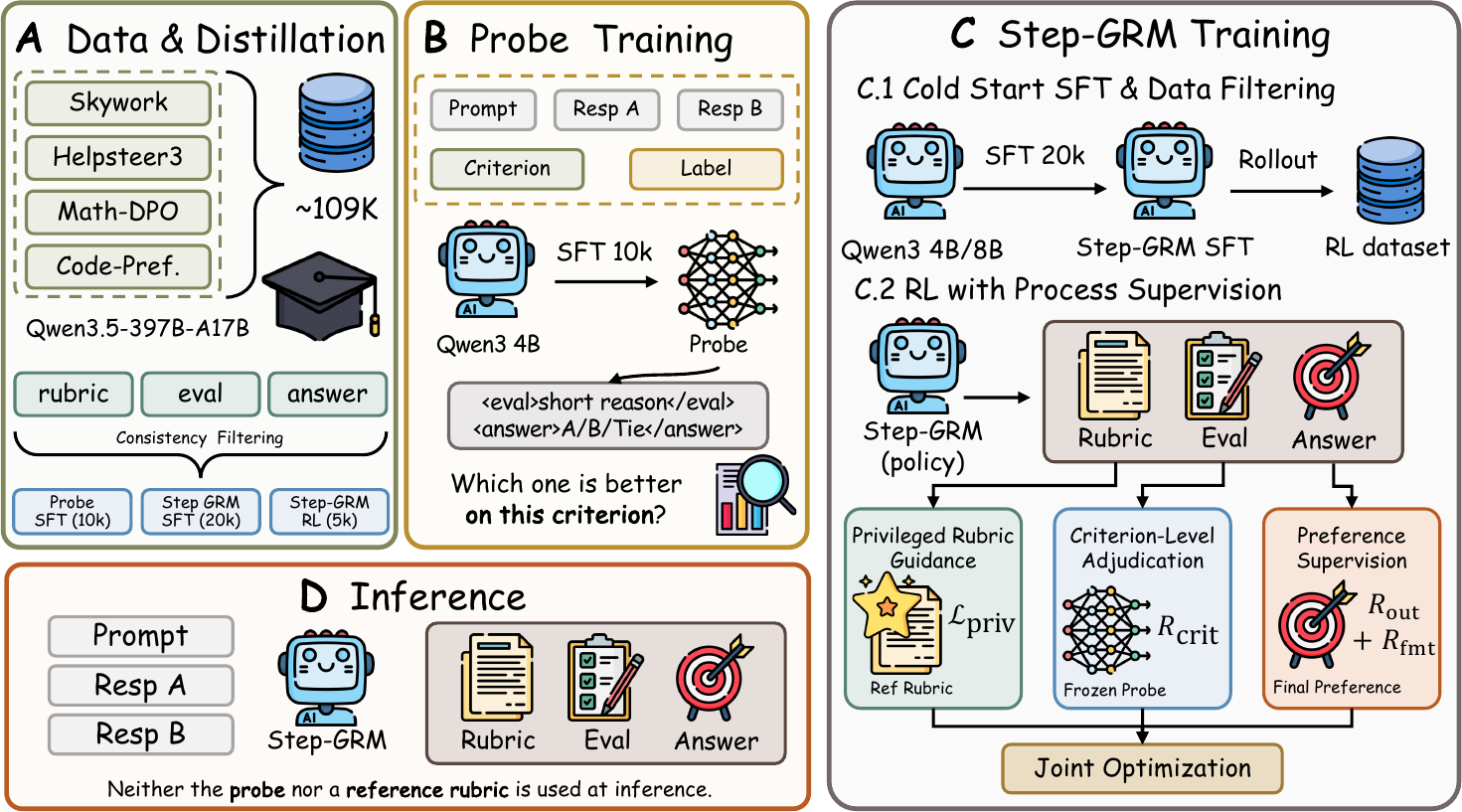}
\caption{Overview of \methodName{}. (A) Structured preference data are constructed and partitioned for probe training, cold-start SFT, and RL. (B) A single-criterion probe is trained to independently judge response pairs under individual criteria and is frozen thereafter. (C) \methodName{} is initialized with SFT and subsequently optimized with process supervision: privileged rubric self-distillation $\mathcal{L}_{\mathrm{priv}}$ guides the evaluation structure, the frozen probe provides criterion-level reward $R_{\mathrm{crit}}$, and outcome and format rewards $R_{\mathrm{out}}$ and $R_{\mathrm{fmt}}$ supervise the final preference and output structure, respectively. (D) At inference time, only \methodName{} is retained, constructing its own rubric, producing criterion-level judgments, and predicting the final preference without reference rubrics or the probe.}
\label{fig:method}
\vspace{-3mm}
\end{figure*}

\methodName{} trains a generative reward model to produce a structured evaluation while directly supervising important decisions within that evaluation. Given an instruction $x$ and two candidate responses $y_A$ and $y_B$, the model generates a rubric $\mathcal{R}=\{r_i\}_{i=1}^{m}$ of instance-specific criteria, a criterion-level judgment in the form of a local verdict $\ell_i\in\{A,B,T\}$ for each criterion, where $T$ denotes a tie, and a final preference $v\in\{A,B\}$. Each criterion describes a preference-agnostic property relevant to the current comparison, and each local verdict compares the two responses specifically under that criterion. Allowing criterion-level ties avoids forcing every criterion to provide evidence for one side when the responses are effectively indistinguishable under that criterion. Figure~\ref{fig:method} provides an overview of the \methodName{} framework.

We initialize the model with supervised fine-tuning on structured teacher evaluation traces containing task-specific criteria, aligned criterion-level judgments, and a final preference. The same traces are decomposed into single-criterion examples for training the single-criterion probe described in Section~\ref{sec:probe}. The main contribution of \methodName{} lies in the subsequent on-policy stage, where the generated evaluation is supervised through two complementary signals: privileged rubric guidance and criterion-level judgment.

\subsection{Privileged Rubric Guidance via Self-Distillation}
\label{sec:rubric-guidance}

The final preference label indicates which response should be selected, but provides little guidance about how the comparison should be organized. Reference rubrics provide complementary information by identifying criteria that are relevant to the current instance. We therefore treat a high-quality reference rubric $\mathcal{R}^{\star}$ as \textbf{privileged information} available only during training and distill its guidance into a policy that remains reference-free at inference time.

Let $c=(x,y_A,y_B)$ denote the standard comparison context. The student (i.e., the policy $\pi_\theta$) first samples an evaluation trajectory $z=(z_1,\ldots,z_L)\sim\pi_\theta(\cdot\mid c)$ without observing the reference rubric. We then construct a privileged context $c^{+}=(x,y_A,y_B,\mathcal{R}^{\star})$ and use the same policy to re-evaluate the student-generated trajectory under this additional information. The privileged branch therefore acts as a self-teacher: it does not produce a separate target completion, but instead assesses the decisions actually made by the current policy when given access to a high-quality evaluation structure.

Concretely, we follow the privileged-context, gap-gated self-distillation formulation of SDAR~\citep{lu2026self}. At token $t$, let
$s_t=(c,z_{<t})$ and $s_t^{+}=(c^{+},z_{<t})$
denote the student and privileged contexts. We compute the detached teacher--student log-probability gap

\[
\Delta_t = \operatorname{sg}\left[ \log \pi_\theta(z_t\mid s_t^{+}) - \log \pi_\theta(z_t\mid s_t) \right],
\]

and convert it into a bounded token-level weight

\[
g_t = \operatorname{sg}\left[ \sigma(\beta\Delta_t) \right],
\]

where $\sigma$ is the logistic sigmoid and $\beta$ controls the sharpness of the gate. The privileged self-distillation objective is

\[
\begin{aligned}
\mathcal{L}_{\mathrm{priv}} &= \operatorname{Agg}_{t}\Bigl[\, g_t \big( \log \pi_\theta(z_t\mid s_t^{+}) \\
& \qquad - \log \pi_\theta(z_t\mid s_t) \big) \Bigr],
\end{aligned}
\]

Here $\operatorname{Agg}_{t}$ denotes averaging over all trajectory tokens.

The likelihood gap measures whether access to the reference rubric provides stronger support for a token already produced by the student. The gate consequently emphasizes student decisions endorsed by the privileged view while attenuating those for which the additional rubric information provides weaker support.

This supervision primarily shapes the \textbf{structure of the evaluation} but does not verify that each criterion-level conclusion is locally supported. We therefore introduce a second signal targeting criterion-level judgments.

\subsection{Criterion-Level Judgment with a Frozen Probe}
\label{sec:probe}

A meaningful criterion does not guarantee a reliable judgment under that criterion. In particular, a model can generate reasonable criteria while assigning their local verdicts in a way that simply mirrors its final preference. To directly supervise these local verdicts, we train a separate \textbf{single-criterion probe}.

\paragraph{Probe Construction.}
We derive the probe training data from the structured teacher evaluation traces used for model initialization. Suppose a teacher trace contains $m'$ criteria and their corresponding local verdicts,
$(r_1,\ell_1),\ldots,(r_{m'},\ell_{m'})$.
We decompose this trace into $m'$ independent criterion-level examples, $(x,y_A,y_B,r_i)\mapsto\ell_i$ with $\ell_i\in\{A,B,T\}$. Each example therefore asks a substantially narrower question than global preference prediction: \emph{given this criterion alone, how do the two responses compare?} We train a separate model $q_\phi(\ell\mid x,y_A,y_B,r)$ with $\ell\in\{A,B,T\}$ on this three-way prediction problem. After training, its parameters $\phi$ are frozen throughout policy optimization. The complete probe prompt is provided in Appendix~\ref{app:prompts} (Figure~\ref{fig:prompt_probe}).

The single-criterion input restriction is deliberate. The probe does not observe the remaining rubric, the student's other criterion-level verdicts, or the student's final preference. Freezing the probe additionally prevents the probe and the policy from co-adapting during reinforcement learning, yielding a fixed local supervision signal as the policy evolves.

\paragraph{Criterion-Level Supervision.}
During on-policy training, the generative reward model constructs its own rubric and assigns local verdicts. Every generated criterion $r_i$ is independently submitted to the frozen probe,

\[
\ell_i^{p} = \argmax_{\ell\in\{A,B,T\}} q_\phi\left( \ell\mid x,y_A,y_B,r_i \right).
\]

We define the criterion-level reward as the average agreement between the policy's local verdicts and the probe predictions,

\[
R_{\mathrm{crit}} = \frac{1}{m} \sum_{i=1}^{m} \mathbbm{1}\!\left[ \ell_i=\ell_i^{p} \right].
\]

Crucially, $R_{\mathrm{crit}}$ is computed without using the student's final preference. Consider a policy that ultimately selects response $A$ and assigns verdict $A$ to every generated criterion. Such a trace can obtain full outcome reward whenever $A$ matches the dataset preference label, even if some of its local claims are unsupported. Under $R_{\mathrm{crit}}$, however, it receives no credit for a criterion whenever the probe predicts $B$ or a tie. Criterion-level supervision therefore discourages intermediate judgments from simply echoing the final preference and directly targets the post-hoc rationalization failure.

Moreover, the probe judges criteria generated by the \textbf{current policy} rather than only the teacher criteria it was trained on, providing local feedback for newly generated criteria without per-rollout annotations.


\subsection{Joint Optimization}
\label{sec:joint-opt}

We combine the two forms of process supervision with the conventional preference objective. For a generated evaluation with final preference $v$, we define the outcome reward as

\[
R_{\mathrm{out}} = \mathbbm{1}\!\left[ v=v^{\star} \right],
\]

where $v^{\star}$ is the dataset preference label. We additionally use a lightweight format reward $R_{\mathrm{fmt}}$ to encourage well-formed structured outputs in which criteria and criterion-level judgments remain aligned. The reinforcement-learning reward is

\[
R_{\mathrm{RL}} = \lambda_{\mathrm{out}}R_{\mathrm{out}} + \lambda_{\mathrm{crit}}R_{\mathrm{crit}} + \lambda_{\mathrm{fmt}}R_{\mathrm{fmt}},
\]

where $\lambda_{\mathrm{out}}$, $\lambda_{\mathrm{crit}}$, and $\lambda_{\mathrm{fmt}}$ are non-negative weighting coefficients.

We optimize this reward using Group Relative Policy Optimization (GRPO)~\citep{shao2024deepseekmath,guo2025deepseek}. For each comparison, multiple structured evaluations are sampled from the current policy, and their relative rewards determine the policy-gradient advantages. The criterion-level reward is computed from the criteria appearing in these student-generated rollouts.

Privileged rubric self-distillation is optimized jointly as an auxiliary objective:

\[
\mathcal{L}(\theta) = \mathcal{L}_{\mathrm{GRPO}}\!\left( \theta; R_{\mathrm{RL}} \right) + \lambda_{\mathrm{priv}} \, \mathcal{L}_{\mathrm{priv}}(\theta),
\]

where $\lambda_{\mathrm{priv}}$ controls the strength of privileged self-distillation.

The three semantic signals address complementary limitations of outcome-only training. $R_{\mathrm{out}}$ determines whether the model assigns the correct final preference. $\mathcal{L}_{\mathrm{priv}}$ provides dense guidance about how the evaluation should be structured when informative reference criteria are available. $R_{\mathrm{crit}}$ directly constrains whether the local judgments made within that structure are supported by their corresponding criteria.

All additional supervision is training-only: at inference time, the model receives only the instruction and the two candidate responses, constructs its own rubric, judges each criterion, and predicts the final preference, without the reference rubric, the privileged self-teacher, or the single-criterion probe.

\begin{table*}[!t]
\caption{
Main results on \textsc{RewardBench} and \textsc{RM-Bench} across domains. \textbf{Bold} and \underline{underlined} numbers indicate the best and second-best results in each column. ``Avg.'' is the mean of the two overall benchmark scores. $^{\star}$ indicates results reported in the original work, with per-domain scores unavailable.
}
\label{tab:main_results}
\centering
\footnotesize
\resizebox{\textwidth}{!}{%
\begin{tabular}{@{}l ccccc ccccc c@{}}
\toprule
\multirow{2}{*}{\textbf{Models}} & \multicolumn{5}{c}{\textbf{RewardBench}} & \multicolumn{5}{c}{\textbf{RM-Bench}} & \multirow{2}{*}{\textbf{Avg.}} \\
\cmidrule(lr){2-6} \cmidrule(lr){7-11}
& \textbf{Chat} & \textbf{Chat-Hard} & \textbf{Safety} & \textbf{Reason.} & \textbf{Overall} & \textbf{Chat} & \textbf{Math} & \textbf{Code} & \textbf{Safety} & \textbf{Overall} & \\
\midrule
\multicolumn{12}{c}{\textit{\textbf{LLM-as-a-Judge}}} \\
\midrule
\texttt{Claude-3-5-Sonnet-20240620} & 96.4 & 74.0 & 81.6 & 84.7 & 84.2 & 62.5 & 62.6 & 54.4 & 64.4 & 61.0 & 72.6 \\
\texttt{Llama-3.1-70B-Instruct} & 97.2 & 70.2 & 82.8 & 86.0 & 84.0 & 64.3 & 67.3 & 47.5 & 83.0 & 65.5 & 74.8 \\
\texttt{GPT-4o-0806} & 96.1 & 76.1 & 86.6 & 88.1 & 86.7 & 67.2 & 67.5 & 63.6 & 91.7 & 72.5 & 79.6 \\
\midrule
\multicolumn{12}{c}{\textit{\textbf{Scalar Reward Models}}} \\
\midrule
\texttt{SteerLM-RM-70B} & 91.3 & 80.3 & \textbf{92.8} & 90.6 & 88.8 & 56.4 & 53.0 & 49.3 & 51.2 & 52.5 & 70.7 \\
\texttt{Eurus-RM-7B} & 98.0 & 65.6 & 81.4 & 86.3 & 82.8 & 59.9 & 60.2 & 56.9 & 86.5 & 65.9 & 74.4 \\
\texttt{InternLM2-7B-Reward} & \textbf{99.2} & 69.5 & 87.2 & 94.5 & 87.6 & 61.7 & 71.4 & 49.7 & 85.5 & 67.1 & 77.4 \\
\texttt{ArmoRM-Llama3-8B-v0.1} & 96.9 & 76.8 & 90.5 & \textbf{97.3} & \underline{90.4} & 67.8 & 57.5 & 53.1 & 92.4 & 67.7 & 79.1 \\
\texttt{InternLM2-20B-Reward} & \underline{98.9} & 76.5 & 89.5 & 95.8 & 90.2 & 63.1 & 66.8 & 56.7 & 86.5 & 68.3 & 79.3 \\
\texttt{Skywork-Reward-Gemma-2-27B} & 95.8 & \textbf{91.4} & \underline{91.9} & \underline{96.1} & \textbf{93.8} & 69.5 & 54.7 & 53.2 & 91.9 & 67.3 & \underline{80.6} \\
\midrule
\multicolumn{12}{c}{\textit{\textbf{Generative Reward Models}}} \\
\midrule
\texttt{RubricARM-Qwen3-8B} & 90.8 & 81.0 & 59.3 & 82.2 & 78.3 & \underline{69.8} & 69.2 & 63.4 & 62.3 & 66.2 & 72.2 \\
\texttt{RubricARROW-Qwen3-8B} & 90.2 & 81.1 & 63.8 & 79.1 & 78.6 & 67.7 & 71.0 & \textbf{66.6} & 71.4 & 69.2 & 73.9 \\
\texttt{RubricRM-Qwen3-8B} & 92.2 & 76.2 & 83.9 & 78.4 & 82.7 & 61.6 & 69.9 & 59.2 & 89.2 & 70.0 & 76.3 \\
\texttt{RM-R1-DeepSeek-Distilled-7B} & 88.9 & 66.2 & 78.4 & 87.0 & 80.1 & 64.0 & \underline{83.9} & 56.2 & 85.3 & 72.4 & 76.3 \\
\texttt{RM-R1-Qwen2.5-7B-Instruct} & 94.1 & 74.6 & 85.2 & 86.7 & 85.2 & 66.6 & 67.0 & 54.6 & 92.6 & 70.2 & 77.7 \\
\texttt{FARE-Qwen3-8B} & 96.5 & 70.7 & 85.7 & 89.2 & 85.5 & 62.0 & 71.6 & 59.2 & 89.4 & 70.5 & 78.0 \\
\texttt{JudgeLRM-Qwen3-8B} & 96.6 & 60.5 & 77.8 & 93.7 & 82.2 & 54.2 & \textbf{86.4} & \underline{65.5} & 91.3 & \underline{74.4} & 78.3 \\
\texttt{J1-Llama-3.1-8B}$^{\star}$ & -- & -- & -- & -- & 85.7 & -- & -- & -- & -- & 73.4 & 79.6 \\
\midrule
\multicolumn{12}{c}{\textit{\textbf{Ours}}} \\
\midrule
\rowcolor{purple!10}
\methodName{}-Qwen3-4B & 91.1 & 82.3 & 87.6 & 80.0 & 85.3 & 69.6 & 71.9 & 61.5 & \underline{92.8} & 74.0 & 79.7 \\
\rowcolor{purple!10}
\methodName{}-Qwen3-8B & 92.7 & \underline{84.8} & 87.7 & 83.4 & 87.2 & \textbf{71.2} & 73.4 & 64.2 & \textbf{93.9} & \textbf{75.7} & \textbf{81.5} \\
\bottomrule
\end{tabular}
}
\end{table*}

\section{Experiments}
\label{sec:experiments}

We evaluate \methodName{} to answer four core research questions:
\begin{itemize}[leftmargin=*,noitemsep,topsep=1pt]
    \item \textbf{RQ1 (Preference Discrimination):} Does supervising the evaluation process improve overall pairwise reward modeling performance across domains? (\S\ref{sec:exp-rq1})
    \item \textbf{RQ2 (Process Reliability and Rationalization):} Does criterion-level judgment supervision mitigate post-hoc rationalization and produce reliable intermediate judgments? (\S\ref{sec:exp-rq2})
    \item \textbf{RQ3 (Downstream Policy Alignment):} Do the learned reward signals translate to superior downstream policy optimization? (\S\ref{sec:exp-rq3})
    \item \textbf{RQ4 (Ablation and Mechanism Analysis):} How do the individual supervision components and architectural designs contribute to the model? (\S\ref{sec:exp-rq4})
\end{itemize}

\subsection{Experimental Setup}
\label{sec:setup}

\paragraph{Training Data.}
We collect approximately 109K preference pairs from Skywork Reward Preference 80K~\citep{liu2026skywork}, \textsc{HelpSteer3}~\citep{wang2025helpsteer3}, Math-DPO-10K~\citep{lai2024step}, and Code-Preference-Pairs~\citep{lai2026r}, and use \texttt{Qwen3.5-397B-A17B} to annotate each pair with a structured evaluation, keeping only examples whose predicted preference matches the dataset preference label. We use 20K examples for supervised fine-tuning and 10K for training the single-criterion probe. After SFT, we perform 8 rollouts per remaining example, keep only comparisons whose preference the SFT policy predicts correctly in 2--6 out of 8 rollouts, and sample 5K for RL, focusing optimization on non-trivial comparisons near the policy's decision boundary. A detailed description of the four data sources is provided in Appendix~\ref{appen_sub:training_data}.

\paragraph{Implementation Details.}
We use \texttt{Qwen3-4B} and \texttt{Qwen3-8B} as reward model backbones and \texttt{Qwen3-4B} as the single-criterion probe, all fully fine-tuned. All models are first fine-tuned with SFT for one epoch with learning rate $1\times10^{-5}$ and global batch size 64 and next optimize with RL for 500 steps using 8 rollouts per prompt, learning rate $1\times10^{-6}$, and maximum completion length 2,048, with $\lambda_{\mathrm{fmt}}=0.2$, $\lambda_{\mathrm{out}}=1.0$, $\lambda_{\mathrm{crit}}=1.0$, $\lambda_{\mathrm{priv}}=0.1$, and $\beta=5.0$. All experiments are conducted on 16 NVIDIA A100 GPUs. The full training configuration is described in Appendix~\ref{appen_sub:impl}.

\paragraph{Evaluation Benchmarks and Metrics.}
We evaluate \methodName{} as a pairwise reward model on two widely used reward-modeling benchmarks: \textsc{RewardBench}~\citep{lambert2025rewardbench} and \textsc{RM-Bench}~\citep{liu2025rm}. \textsc{RewardBench} measures preference discrimination across \texttt{Chat}, \texttt{Chat-Hard}, \texttt{Safety}, and \texttt{Reasoning}. \textsc{RM-Bench} probes sensitivity to response content while penalizing reward models that can be swayed by stylistic bias, and reports \texttt{Chat}, \texttt{Math}, \texttt{Code}, and \texttt{Safety} separately. We follow the official splits and scoring rules of each benchmark, report accuracy (\%) per domain, and average the two overall scores. Detailed benchmark descriptions and the evaluation baselines are provided in Appendices~\ref{appen_sub:benchmark} and~\ref{appen_sub:baselines}.

%
%

\subsection{Preference Discrimination (RQ1)}
\label{sec:exp-rq1}

Table~\ref{tab:main_results} compares \methodName{} against representative LLM judges, scalar reward models, and generative reward models on \textsc{RewardBench} and \textsc{RM-Bench}. Supervising both rubric construction and criterion-level judgment yields consistent gains across model scales: the 4B variant is competitive with larger 8B generative baselines, while the 8B variant achieves the highest overall average score across the two benchmarks. Although certain baseline models perform strongly on specific individual subsets, our model discriminates more consistently across the full suite of evaluations, indicating that supervising intermediate evaluation steps helps generative reward models produce more dependable pairwise preferences across varied settings.


\begin{table}[t]
\centering
\small
\caption{
Evaluation of process reliability on the \textsc{HelpSteer3} validation split.
Process metrics are computed against human-derived reference evaluations.
Each ``w/o'' row removes a single supervision signal from \methodName{}, whereas ``GRPO Outcome-Only'' is trained with the outcome reward alone.
}
\label{tab:human_process}

\resizebox{\linewidth}{!}{%
\begin{tabular}{lccc}
\toprule
\textbf{Model}
& \textbf{Out.\ Acc.} $\uparrow$
& \textbf{Proc.\ Cons.} $\uparrow$
& \textbf{Local Acc.} $\uparrow$ \\
\midrule
Cold-Start SFT
& 72.75 & 30.18 & 58.62 \\

GRPO Outcome-Only
& 74.93 & 27.43 & 54.28 \\

w/o $\mathcal{L}_{\mathrm{priv}}$
& 76.15 & 28.76 & 63.37 \\

w/o $R_{\mathrm{crit}}$
& 75.56 & 32.47 & 56.45 \\

\rowcolor{purple!10}
\textbf{\methodName{}}
& \textbf{76.68}
& \textbf{35.25}
& \textbf{65.75} \\
\bottomrule
\end{tabular}%
}
\end{table}

\subsection{Process Reliability and Rationalization (RQ2)}
\label{sec:exp-rq2}



To verify whether the evaluation process learned by \methodName{} genuinely aligns with human judgment rather than merely overfitting to the auxiliary probe, we conduct evaluations on the official validation split of \textsc{HelpSteer3}~\citep{wang2025helpsteer3}. Following the evaluation protocol of \citet{wang2026outcome}, we employ \texttt{Qwen3.5-Plus} both to structure the rich human response-level feedback into reference evaluations and to compute the intermediate process metrics. Table~\ref{tab:human_process} reports Outcome Accuracy (the fraction of examples whose final preference matches the human reference) alongside two process-level metrics: Process Consistency (semantic alignment between generated criteria and human references) and Local Accuracy (verdict correctness on matched criteria).

The empirical results show that outcome-only supervision can improve final preference accuracy while degrading both Process Consistency and Local Accuracy, revealing a clear rationalization drift in the intermediate evaluation process. In contrast, \methodName{} consistently improves both process metrics while maintaining strong final preference accuracy. Ablation results further indicate that privileged rubric guidance ($\mathcal{L}_{\mathrm{priv}}$) is essential for identifying relevant criteria, whereas criterion-level supervision ($R_{\mathrm{crit}}$) directly improves the reliability of local verdicts and mitigates post-hoc rationalization. Together, these observations demonstrate that supervising both rubric structure and local verdicts better aligns the intermediate evaluation process with human-derived reference evaluations.

\begin{table}[t]
\centering
\caption{
Comparison of different strategies applied to \texttt{Qwen2.5-7B-Instruct} on \textsc{Arena-Hard} and \textsc{AlpacaEval}. 
\textbf{WC} denotes the win rate (against \texttt{GPT-4-0314} on \textsc{Arena-Hard}), and \textbf{LC} denotes the length-controlled win rate on \textsc{AlpacaEval}. 
\textbf{Bold} and \underline{underlined} numbers indicate the best and second-best results in each column.
}
\label{tab:alpaca}
\resizebox{\linewidth}{!}{%
\arrayrulecolor{black!50}
\begin{tabular}{l|c|cc|c}
\toprule
\multirow{2}{*}{\textbf{Model}} 
& \multicolumn{1}{c}{\textbf{Arena-Hard}} 
& \multicolumn{2}{c|}{\textbf{AlpacaEval}} 
& \multirow{2}{*}{\textbf{Avg.}} \\ 
\cmidrule(lr){2-2} \cmidrule(lr){3-4}
& \textbf{WC} 
& \textbf{WC} & \textbf{LC} &  \\ 
\midrule
\texttt{GPT-4-0314} & 50.0 & 22.1 & 35.3 & 35.8 \\ 
\midrule
\texttt{Qwen2.5-7B-Instruct} & 51.3 & 33.5 & 36.2 & 40.3 \\
+ SFT (Distilled)           & 32.6 & 36.1 & 33.3 & 34.0 \\
+ DPO (via \texttt{Skywork-RM})         & \underline{55.1} & 44.8 & \underline{41.5} & \underline{47.1} \\
+ DPO (via \texttt{ArmoRM})           & 50.8 & 37.6 & 38.1 & 42.2 \\
+ DPO (via \texttt{UltraFeedback})        & 52.8 & 33.7 & 38.7 & 41.7 \\
+ DPO (via LLM-as-a-Judge)    & 51.0 & 28.8 & 33.4 & 37.7 \\
+ DPO (via RLCF)             & 54.6 & 36.2 & 37.1 & 42.6 \\ 
+ DPO (via \texttt{RubricRM})          & 52.9 & \underline{47.0} & 41.3 & \underline{47.1} \\
\midrule
\rowcolor{purple!10}
+ DPO (via \methodName{}-Qwen3-4B)            & 53.3 & 46.2 & 41.2 & 46.9 \\ 
\rowcolor{purple!10}
+ DPO (via \methodName{}-Qwen3-8B)            & \textbf{56.6} & \textbf{47.7} & \textbf{42.1} & \textbf{48.8} \\
\bottomrule
\end{tabular}%
}
\end{table}

\begin{figure*}[t]
    \centering
    \includegraphics[width=0.485\textwidth]{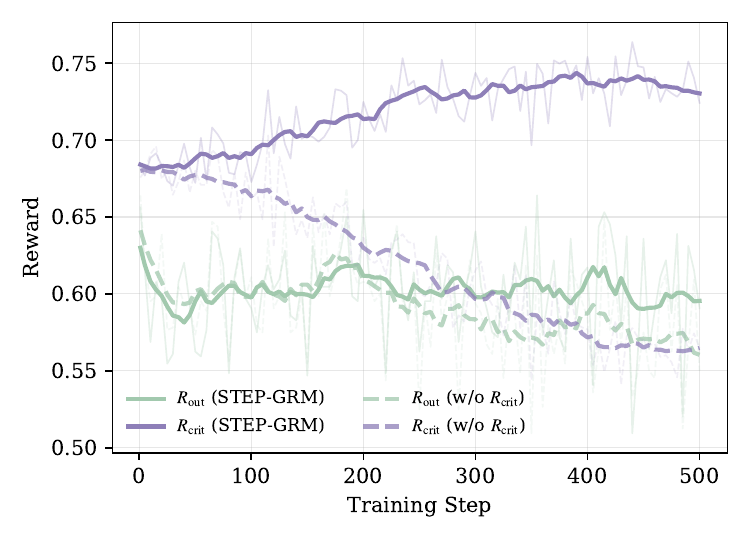}
    \hfill
    \includegraphics[width=0.485\textwidth]{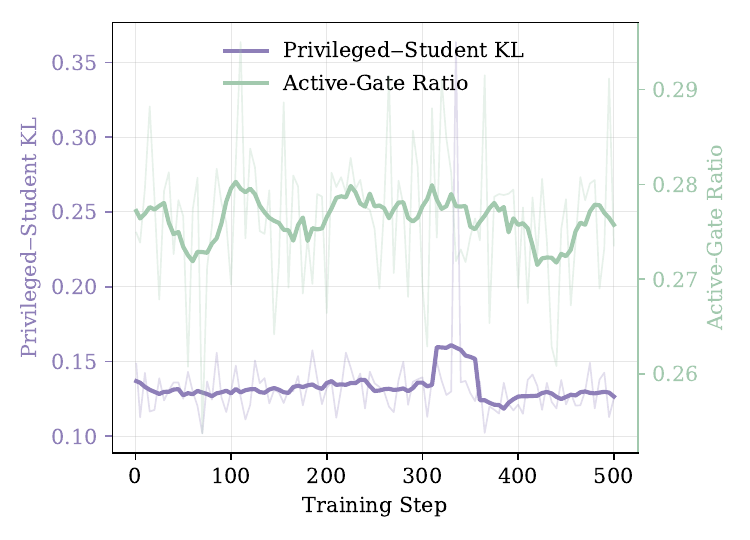}
    \caption{\textbf{Training dynamics of process supervision.}
    \textbf{Left:} Removing $R_{\mathrm{crit}}$ preserves outcome reward growth but leads to a sharp collapse in criterion-level agreement, illustrating post-hoc rationalization drift.
    \textbf{Right:} Privileged guidance maintains a stable KL divergence with a sparse gate activation ratio of $\approx 27$--$28\%$.}
    \label{fig:training-dynamics}
    \vspace{-2mm}
\end{figure*}

\subsection{Downstream Policy Alignment (RQ3)}
\label{sec:exp-rq3}

To examine whether the learned reward signals benefit downstream generation, we apply different reward models to guide offline preference optimization via DPO on \texttt{Qwen2.5-7B-Instruct}, following the experimental configuration of \citet{viswanathan2026checklists}. As reported in Table~\ref{tab:alpaca}, policies aligned with our reward models achieve competitive or higher win rates on \textsc{Arena-Hard} and \textsc{AlpacaEval} compared to those trained with baseline reward models. In particular, our 8B reward model achieves the highest overall average win rate, outperforming both scalar and rubric-based baselines. Importantly, the gains on \textsc{AlpacaEval} remain consistent under the length-controlled metric, suggesting that the improvement reflects better response quality rather than length bias. These results confirm that process-supervised reward modeling provides practical and reliable preference feedback for policy post-training.


\subsection{Ablation and Mechanism Analysis (RQ4)}
\label{sec:exp-rq4}

\begin{table}[t]
\centering
\small
\caption{Ablation results on training components and probe design with the \texttt{Qwen3-4B} backbone, evaluated on \textsc{RewardBench} and \textsc{RM-Bench}.}
\label{tab:ablations-combined}
\resizebox{\linewidth}{!}{%
\begin{tabular}{l ccc}
\toprule
\textbf{Configuration} & \textbf{RewardBench} & \textbf{RM-Bench} & \textbf{Avg.} \\
\midrule
\rowcolor{purple!10}
\textbf{\methodName{} (Full)} & \textbf{85.3} & \textbf{74.0} & \textbf{79.7} \\
\midrule
\multicolumn{4}{l}{\textit{Training Components}} \\
\quad w/o $R_{\mathrm{crit}}$ & 83.7 & 73.5 & 78.6 \\
\quad w/o $\mathcal{L}_{\mathrm{priv}}$ & 83.9 & 73.9 & 78.9 \\
\quad w/o Gap Gate & 84.3 & 73.8 & 79.0 \\
\quad Cold-Start SFT & 83.1 & 72.1 & 77.6 \\
\quad GRPO Outcome-Only & 83.5 & 73.2 & 78.3 \\
\midrule
\multicolumn{4}{l}{\textit{Probe Design \& Verdict Space}} \\
\quad Forced Binary (w/o Tie) & 85.1 & 73.7 & 79.4 \\
\quad Probe SFT + GRPO & 85.1 & 73.9 & 79.5 \\
\quad \texttt{Qwen3-4B} (Zero-Shot) & 83.7 & 73.2 & 78.5 \\
\quad \texttt{Qwen3-8B} (Trained) & 85.4 & 74.0 & 79.7 \\
\quad \texttt{Qwen3-32B} (Zero-Shot) & 84.3 & 73.5 & 78.9 \\
\bottomrule
\end{tabular}%
}
\end{table}

\paragraph{Training Components.}
Table~\ref{tab:ablations-combined} isolates the contribution of each supervision signal in \methodName{}. Removing criterion-level supervision ($R_{\mathrm{crit}}$) results in the largest drop among the process-supervision ablations, highlighting the importance of directly constraining local verdicts. Disabling privileged rubric self-distillation (w/o $\mathcal{L}_{\mathrm{priv}}$) also degrades performance, confirming that reference rubrics offer beneficial structural priors during training. In addition, removing the gap-gated mechanism leads to lower scores, suggesting that selective distillation is more effective than uniformly applying privileged guidance. Overall, all process-supervised variants consistently outperform the outcome-only baseline, underscoring the complementary benefits of structural and local supervision.

\paragraph{Probe Design and Verdict Space.}
Table~\ref{tab:ablations-combined} investigates key design choices in the criterion-level judgment mechanism. Restricting criterion-level verdicts to a forced binary choice without ties degrades performance, showing that allowing intermediate ties provides a calibrated option that prevents the model from inventing spurious differences. For probe training, task-specific alignment is essential: an unaligned zero-shot probe performs poorly, whereas further RL training on top of SFT brings no noticeable gain. Furthermore, scaling the trained probe from 4B to 8B yields only a marginal difference, while an unaligned 32B zero-shot model underperforms the aligned 4B probe. This suggests that reliable local supervision relies primarily on targeted task alignment rather than evaluator scale.

\paragraph{Training Dynamics.}
Figure~\ref{fig:training-dynamics} illustrates the internal optimization dynamics during reinforcement learning. When criterion-level reward is withheld, the policy steadily improves its final outcome reward while the monitored criterion agreement drops sharply, reflecting that the model learns to bend intermediate rationales to favor the final preference. Incorporating $R_{\mathrm{crit}}$ successfully stabilizes criterion agreement throughout training, directly preventing this rationalization drift. Concurrently, the privileged self-distillation dynamics exhibit a persistent KL divergence and a gate activation ratio of around $27$--$28\%$, confirming that the reference rubric provides targeted structural guidance without indiscriminately overriding the student policy.

\section{Related Work}
\label{sec:related_work}

\paragraph{Reward Modeling.}
Reward models provide learned proxies for human preferences and have become a central component of language-model alignment. Early preference-learning approaches learned scalar reward functions from pairwise comparisons~\citep{christiano2017deep}, and this paradigm was subsequently adapted to language generation and RLHF~\citep{ziegler2019fine,stiennon2020learning,ouyang2022training,bai2022constitutional}. Most conventional reward models remain discriminative, directly predicting a scalar score or preference from candidate responses, with later work extending this formulation to multi-dimensional and interpretable preference modeling~\citep{wang2024interpretable}. In parallel, LLM-as-a-Judge has emerged as a flexible alternative for model-based evaluation~\citep{zheng2023judging}, motivating methods that use language models themselves to generate feedback or rewards~\citep{yuan2024self,wang2024self}. More recent generative reward models explicitly produce critiques, rationales, or reasoning traces before making reward or preference predictions~\citep{mahan2024generative,ye2024beyond,ankner2024critique,zhang2025generative}, while reasoning-oriented approaches further optimize the evaluation reasoning process through distillation or reinforcement learning~\citep{chen2026rm,huang2026think}. Recent studies have also highlighted that correct preference predictions can coexist with unreliable evaluation rationales~\citep{wang2026outcome,lai2026r}. \methodName{} builds on this generative line of reward modeling, with a particular focus on improving the reliability of its structured intermediate evaluation process.

\paragraph{Rubric-Based Evaluation and Reward Modeling.}
A complementary line of work structures evaluation around explicit criteria or rubrics rather than holistic judgments. Early LLM-based evaluators introduced task-specific evaluation criteria, fine-grained skill decomposition, and customized scoring rubrics~\citep{liu2023g,ye2024flask,kim2024prometheus,kim2024prometheus2,hashemi2024llm}. Recent work increasingly shifts from fixed, externally specified rubrics toward \emph{instance-specific and automatically generated rubrics} that can serve as structured evaluation or reward signals. Rubrics as Rewards directly uses instance-specific rubrics for reinforcement learning~\citep{gunjal2026rubrics}, while OpenRubrics develops scalable synthetic rubric generation and trains rubric-based reward models~\citep{liu2026openrubrics}. Subsequent studies investigate rubric design for reward misspecification~\citep{zhang2026chasing}, online rubric elicitation as the policy evolves~\citep{rezaei2025online}, and joint or alternating optimization of rubric generators and rubric-conditioned judges~\citep{xu2026alternating,jiang2026rubric}. Concurrent work also explores automatically generating and refining dynamic evaluation rubrics through meta-judge feedback~\citep{wang2026generating}. Together, these studies establish rubrics as an increasingly important interface for decomposing open-ended evaluation into explicit, interpretable criteria. \methodName{} follows this direction while treating both rubric construction and criterion-level judgment as intermediate components of the evaluation process that receive dedicated training-time supervision.

\section{Conclusion}
\label{sec:conclusion}

We present \methodName{}, a framework that supervises the intermediate evaluation process of generative reward models to address post-hoc rationalization under outcome-only training. By combining privileged rubric self-distillation with independent criterion-level judgment via a frozen probe, \methodName{} constrains both rubric construction and local verdicts during reinforcement learning without requiring either component at inference time. Empirical results show that supervising intermediate evaluation steps improves pairwise preference discrimination, aligns intermediate rationales with human references, and provides effective reward signals for downstream policy optimization.

\section*{Limitations}

Our current study leaves several directions for future investigation. First, our experiments primarily focus on Qwen-based backbones at the 4B and 8B scales. Although \methodName{} shows consistent improvements across both model sizes, evaluating the framework on broader model families and substantially larger scales would further establish its generality. Second, our empirical evaluation mainly considers pairwise preference modeling on existing reward-modeling benchmarks and downstream policy optimization with DPO. Extending process-supervised reward modeling to other evaluation formats and alignment algorithms is an interesting direction for future work. Finally, \methodName{} introduces additional training-time computation through privileged self-distillation and criterion-level supervision. Exploring more lightweight supervision mechanisms could further improve training efficiency while preserving the reference-free inference procedure.


\bibliography{custom}

\newpage
\appendix

\section{Experiment Setup}
\label{appen:exp_setups}

\subsection{Benchmarks}
\label{appen_sub:benchmark}

We evaluate pairwise reward quality on two complementary benchmarks, \textsc{RewardBench} and \textsc{RM-Bench}, which probe different aspects of reward modeling.

\paragraph{\textsc{RewardBench}.}
\textsc{RewardBench}~\citep{lambert2025rewardbench} is one of the first dedicated testbeds for reward models, built as triplets that pair each instruction with a preferred response and an inferior response drawn from real-world conversations and post-training data. It comprises four subsets: \texttt{Chat} (350), \texttt{Chat-Hard} (456), \texttt{Safety} (1,431), and \texttt{Reasoning} (740). \texttt{Chat-Hard} emphasizes adversarial instructions where the preferred response is less sycophantic or explicitly refuses, whereas \texttt{Safety} and \texttt{Reasoning} aggregate contrastive pairs from moderation scenarios and math, coding, and instruction-following sources. Following the official protocol, we report per-subset accuracy, i.e., the fraction of triplets for which the chosen response receives the higher reward.

\paragraph{\textsc{RM-Bench}.}
\textsc{RM-Bench}~\citep{liu2025rm} targets two failure modes that \textsc{RewardBench} rarely exposes: insensitivity to subtle content differences and susceptibility to stylistic bias. Each query pairs a stronger response containing a deliberately injected error with a weaker response that is restyled to appear superficially more polished, so that a reward model scores well only if it values substance over style. The benchmark covers four domains, namely \texttt{Chat} (129), \texttt{Math} (529), \texttt{Code} (228), and \texttt{Safety} (441). In addition, every comparison is instantiated at three difficulty levels that control how subtly the error is introduced. We report the official per-domain accuracy averaged over difficulty variants.

\subsection{Training Data}
\label{appen_sub:training_data}

Training comparisons are drawn from four public preference sources that jointly cover general conversation, mathematics, and code.

\textbf{Skywork Reward Preference 80K}~\citep{liu2026skywork} is a large-scale preference collection spanning chat, safety, mathematics, and code. Its constituent sources are curated with a combination of rule-based filtering and human verification, making it a reliable backbone for general-domain reward model training.

\textbf{HelpSteer3}~\citep{wang2025helpsteer3} provides human-annotated preferences over general, STEM-centric, and multilingual tasks, where each response pair is accompanied by a free-form response-level rationale written by professional annotators. Besides serving as a training source, its official validation split is reused for the human-grounded process evaluation in Section~\ref{sec:exp-rq2}.

\textbf{Math-DPO-10K}~\citep{lai2024step} focuses on mathematical reasoning and constructs step-level preferences by pairing correct steps of long chain-of-thought solutions against erroneous alternatives.

\textbf{Code-Preference-Pairs}~\citep{lai2026r} is built by mutating seed programs with deliberately introduced bugs; corrected and broken versions form preference pairs, and surface artifacts such as error comments are removed so that the preference cannot be recovered from stylistic cues.

Following the annotation protocol in Section~\ref{sec:setup}, each retained pair is annotated by \texttt{Qwen3.5-397B-A17B} with a structured evaluation containing task-specific criteria, criterion-level verdicts, and a final preference; examples whose predicted preference contradicts the dataset label are discarded. The resulting pool of approximately 109K pairs is partitioned into 20K comparisons for cold-start SFT, 10K decomposed single-criterion examples for probe training, and 5K boundary comparisons for RL.

\subsection{Baselines}
\label{appen_sub:baselines}

We compare \methodName{} against reward models from three families, following the taxonomy of Table~\ref{tab:main_results}.

\paragraph{LLM-as-a-Judge.}
This family prompts off-the-shelf LLMs to directly compare two candidate responses without any task-specific training~\citep{zheng2023judging}. We include \texttt{GPT-4o-0806}~\citep{hurst2024gpt}, \texttt{Claude-3.5-Sonnet}~\citep{anthropic2024claude}, and \texttt{Llama-3.1-70B-Instruct}~\citep{grattafiori2024llama}. These judges benefit from broad world knowledge, but they apply no structured evaluation protocol and are therefore prone to stylistic shortcuts.

\paragraph{Scalar Reward Models.}
Scalar models map each response to a single score and predict the winner by comparing scores. They are computationally cheap and often strong on in-distribution subsets, yet provide no explicit account of why one response is preferred. We evaluate \texttt{SteerLM-RM-70B}~\citep{wang2024helpsteer}, \texttt{Eurus-RM-7B}~\citep{yuan2025advancing}, \texttt{InternLM2-7B-Reward} and \texttt{InternLM2-20B-Reward}~\citep{cai2024internlm2}, \texttt{ArmoRM-Llama3-8B}~\citep{wang2024interpretable}, and \texttt{Skywork-Reward-Gemma-2-27B}~\citep{liu2024skywork}.

\paragraph{Generative Reward Models.}
Generative reward models produce free-form textual judgments before committing to a final preference. This group includes reasoning-enhanced RMs trained with critique- or RL-based pipelines, such as \texttt{RM-R1}~\citep{chen2026rm}, \texttt{JudgeLRM}~\citep{chen2025judgelrm}, \texttt{J1}~\citep{whitehouse2026j1}, and \texttt{FARE}~\citep{xu2026foundational}. It also covers rubric-based RMs, which anchor their judgments in explicit evaluation criteria, namely \texttt{RubricRM}~\citep{zhang2026chasing}, \texttt{RubricARM}~\citep{jia2026open}, and \texttt{RUBRIC-ARROW}~\citep{jiang2026rubric}.

\paragraph{Downstream Alignment Baselines.}
For the policy optimization experiments in Section~\ref{sec:exp-rq3}, we adopt the DPO configuration of \citet{viswanathan2026checklists} and align \texttt{Qwen2.5-7B-Instruct} with preferences derived from different reward sources: the implicit reward of UltraFeedback~\citep{cui2023ultrafeedback}, scalar RMs (\texttt{Skywork-RM}~\citep{liu2024skywork} and \texttt{ArmoRM}~\citep{wang2024interpretable}), pointwise \texttt{GPT-4o} judgments~\citep{zheng2023judging}, and RLCF~\citep{viswanathan2026checklists}. We also report an SFT variant distilled from teacher responses, and use \texttt{GPT-4-0314}~\citep{achiam2023gpt} as the fixed reference for win-rate computation.

\subsection{Implementation Details}
\label{appen_sub:impl}

All models are fully fine-tuned in bfloat16 with ms-swift~\citep{zhao2025swift}, which provides the pipeline for both the SFT and the RL stage, and training proceeds in three stages. Stage~1 trains the single-criterion probe on 10K decomposed criterion-level examples distilled from the teacher traces; the probe is then frozen and used only to compute the criterion-level reward $R_{\mathrm{crit}}$ during RL. Stage~2 performs the cold-start SFT on 20K evaluation traces to teach the structured evaluation format, serving both as the initialization for policy optimization and as the pure-SFT baseline. Stage~3 optimizes the policy from the Stage-2 checkpoint with GRPO on 5K boundary comparisons, jointly with privileged rubric self-distillation and the outcome, criterion, and format rewards.

Both SFT stages are trained for one epoch with learning rate $1\times10^{-5}$, a cosine schedule with 3\% warmup, global batch size 64, weight decay 0.1, and gradient-norm clipping at 1.0; we use DeepSpeed ZeRO stage 2 for the probe and stage 1 for the judge, together with FlashAttention-2 and gradient checkpointing, and the maximum sequence length is 6,144 for the probe and 8,192 for the judge. The RL stage runs for 500 steps at learning rate $1\times10^{-6}$ with the same schedule and regularization settings, samples 16 independent prompts per step with 8 rollouts each at temperature 1.0, uses no KL regularization, and is optimized with DeepSpeed ZeRO-3, with completions capped at 2,048 tokens within a maximum sequence length of 8,192 (overlong completions are filtered). The reward weights are $\lambda_{\mathrm{fmt}}=0.2$, $\lambda_{\mathrm{out}}=1.0$, and $\lambda_{\mathrm{crit}}=1.0$, and privileged self-distillation uses $\lambda_{\mathrm{priv}}=0.1$ with gate sharpness $\beta=5.0$.

During RL, every criterion generated by the policy is submitted to the frozen probe individually; the probe is served with vLLM and decoded greedily, and its \{A, B, Tie\} predictions define $R_{\mathrm{crit}}$. At inference time, all models are served with vLLM and decoded greedily (temperature 0) with a maximum of 2,048 generated tokens, with the thinking mode disabled to match the output template used during training. To control for position bias, each comparison is also evaluated with the two candidate responses swapped, and we report the average accuracy over the two orders.

\section{Prompts}
\label{app:prompts}

We present the prompts used in this work: the system and user prompts of the \methodName{} judge, the teacher prompt used for privileged rubric self-distillation, and the system prompt of the single-criterion probe.

\begin{figure*}[!t]
\begin{promptbox}[\methodName{} (Judge) System Prompt]
\footnotesize
You are a fair and impartial judge. You are given an instruction and two candidate responses (Response A and Response B). Your job is to decide which response is better by FIRST proposing your own evaluation rubric, then judging against it.

Work in two stages, and output them in exactly the format shown below.

\textbf{1.~Rubric.} Propose a numbered list of concrete, self-contained evaluation criteria. Each criterion must:
\begin{itemize}[leftmargin=*, noitemsep, topsep=2pt]
    \item begin with a tag, either \textbf{[Hard Rule]} or \textbf{[Principle]}: \\
        \textbf{[Hard Rule]}: an objective, verifiable requirement (format, length, required or forbidden content). \\
        \textbf{[Principle]}: a qualitative standard (clarity, depth, correctness, helpfulness).
    \item be phrased abstractly (say ``The response should ...'', never mention ``Response A/B'' inside the criterion text).
    \item end with ``::'' followed by ONE importance label that states which response the criterion favors, based ONLY on this single criterion in isolation: \\
        \textbf{A}: this criterion favors Response A \\
        \textbf{neutral}: both responses do about equally on this criterion (non-discriminative) \\
        \textbf{B}: this criterion favors Response B
    \item Judge each criterion independently: a criterion is neutral whenever both responses satisfy it (or both fail it) to a similar degree. Do not force every criterion to point at the eventual winner.
    \item Each criterion must state a specific, checkable requirement grounded in this particular instruction. Never leave a criterion vague or elliptical (e.g., ``the response should ...'' is invalid).
\end{itemize}

\textbf{2.~Verdict.} Weigh the criteria (Hard Rules gate; Principles break ties) and give the final answer.

Output format, and nothing else (example criteria shown for an imaginary cooking instruction):

\vspace{0.3em}
\begin{verbatim}
<rubrics>
1. [Hard Rule] The response should list exact quantities for every ingredient it mentions. :: A
2. [Principle] The response should explain why each cooking step is necessary. :: neutral
</rubrics>
<answer>A</answer>
\end{verbatim}
\vspace{0.3em}

Use \texttt{<answer>A</answer>} if Response A is better, or \texttt{<answer>B</answer>} if Response B is better.
\end{promptbox}
\caption{\textbf{\methodName{} judge system prompt.} The generative reward model first proposes a rubric of instance-specific criteria, each tagged [Hard Rule] or [Principle] and labelled with the response it favors in isolation, and then aggregates these criterion-level judgments into the final preference.}
\label{fig:prompt_judge}
\end{figure*}

\begin{figure}[!t]
\begin{promptbox}[\methodName{} (Judge) User Prompt]
\footnotesize
\textbf{Instruction:}\\
\{instruction\}

\vspace{0.3em}
\textbf{Response A:}\\
\{response\_a\}

\vspace{0.3em}
\textbf{Response B:}\\
\{response\_b\}

\vspace{0.3em}
Propose your rubric and decide which response is better, following the required format exactly.
\end{promptbox}
\caption{\textbf{\methodName{} judge user prompt.} The instruction and the two candidate responses are filled into the placeholders.}
\label{fig:prompt_judge_user}
\end{figure}

\begin{figure}[!t]
\begin{promptbox}[Teacher User Prompt]
\footnotesize
\textbf{Instruction:}\\
\{instruction\}

\vspace{0.3em}
\textbf{Response A:}\\
\{response\_a\}

\vspace{0.3em}
\textbf{Response B:}\\
\{response\_b\}

\vspace{0.3em}
\textbf{Reference rubric} (expert-provided criteria for this instruction; each criterion has been labelled by a calibrated evaluator with the response it favors, judged on that criterion alone):\\
\{rubric\}

\vspace{0.3em}
Evaluate each reference criterion against the two responses, aggregate the labelled criteria, and give the verdict, following the required format exactly.
\end{promptbox}
\caption{\textbf{Teacher user prompt.} Used for privileged rubric self-distillation: the self-teacher is conditioned on the reference rubric and re-evaluates the two responses, verifying each expert-labelled criterion and aggregating the labelled criteria into the final preference.}
\label{fig:prompt_teacher}
\end{figure}

\begin{figure*}[!t]
\begin{promptbox}[Single-Criterion Probe System Prompt]
\footnotesize
You are a fair and impartial judge. You are given an instruction, two candidate responses (Response 1 and Response 2), and ONE criterion drawn from a rubric. The criterion is tagged \textbf{[Hard Rule]} or \textbf{[Principle]}.

\begin{itemize}[leftmargin=*, noitemsep, topsep=2pt]
    \item \textbf{[Hard Rule]}: an objective, verifiable requirement (e.g., format, length, required or forbidden content).
    \item \textbf{[Principle]}: a qualitative standard (e.g., clarity, depth, helpfulness) that reflects overall quality.
\end{itemize}

Weigh how well each response satisfies THE GIVEN CRITERION ONLY. Judge STRICTLY AND ONLY BY THE GIVEN CRITERION. Do not rely on your own preferences, outside knowledge, or any criterion that is not listed.

Before answering, give a one or two sentence reason for your judgment:
\begin{itemize}[leftmargin=*, noitemsep, topsep=2pt]
    \item quote or paraphrase the specific evidence in each response that matters for this criterion
    \item do not repeat the criterion text, do not list unrelated differences
\end{itemize}

Then output the verdict in exactly this format:

\vspace{0.3em}
\begin{verbatim}
<reason>your concise justification</reason>
<answer>1</answer>   (the criterion favors Response 1)
<answer>2</answer>   (the criterion favors Response 2)
<answer>0</answer>   (both responses satisfy or fail the criterion to a similar degree)
\end{verbatim}
\vspace{0.3em}

No other text after the \texttt{<answer>} tag.
\end{promptbox}
\caption{\textbf{Single-criterion probe system prompt.} The probe judges the two candidate responses under a single criterion in isolation and outputs a reason together with a three-way verdict (A, B, or tie).}
\label{fig:prompt_probe}
\end{figure*}

\end{document}